%% file: main.tex
\documentclass{article} 
\usepackage{iclr2023_conference,times}

\input{math_commands.tex}

\usepackage{hyperref}
\usepackage{url}
\usepackage{graphicx}

\graphicspath{ {./graphics/} }

\title{Exploring a Gradient-based Explainable AI Technique for Time-Series Data: A case study of assessing stroke rehabilitation exercises}

\author{Min Hun Lee \& Yi Jing Choy 
\\
School of Computing and Information Systems\\
Singapore Management University\\
Singapore, 178902, Singapore \\
\texttt{mhlee@smu.edu.sg; yijing.choy.2019@scis.smu.edu.sg}\\
}

\iclrfinalcopy 
\begin{document}

\maketitle

\begin{abstract}
Explainable artificial intelligence (AI) techniques are increasingly being explored to provide insights into why AI and machine learning (ML) models provide a certain outcome in various applications. However, there has been limited exploration of explainable AI techniques on time-series data, especially in the healthcare context. In this paper, we describe a threshold-based method that utilizes a weakly supervised model and a gradient-based explainable AI technique (i.e. saliency map) and explore its feasibility to identify salient frames of time-series data. Using the dataset from 15 post-stroke survivors performing three upper-limb exercises and labels on whether a compensatory motion is observed or not, we implemented a feed-forward neural network model and utilized gradients of each input on model outcomes to identify salient frames that involve compensatory motions. According to the evaluation using frame-level annotations, our approach achieved a recall of 0.96 and an F2-score of 0.91. Our results demonstrated the potential of a gradient-based explainable AI technique (e.g. saliency map) for time-series data, such as highlighting the frames of a video that therapists should focus on reviewing and reducing the efforts on frame-level labeling for model training.
\end{abstract}

\input{iclr2023/contents/intro.tex}
\input{iclr2023/contents/data_preprocessing.tex}
\input{iclr2023/contents/model_training.tex}
\input{iclr2023/contents/understand_saliencymap.tex}

\input{iclr2023/contents/analysing_score.tex}

\subsubsection*{Acknowledgments}
This work has been supported by the Singapore Ministry of Education (MOE) Academic Research Fund (AcRF) Tier 1 grant.

\bibliography{main}
\bibliographystyle{iclr2023_conference}

\newpage
\appendix
\section{Appendix}

\begin{figure}[h]
    \centering
    \includegraphics[scale=0.6]{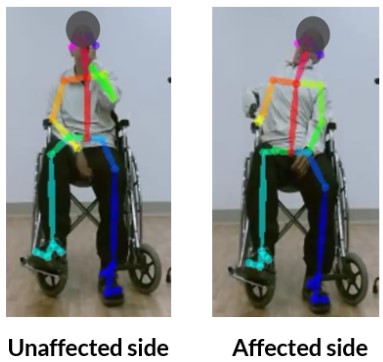}
    \caption{Movement of Unaffected and Affected Side}
    \label{fig:movement-ex}
\end{figure}

\begin{figure}[h]
    \centering
    \includegraphics[scale=0.5]{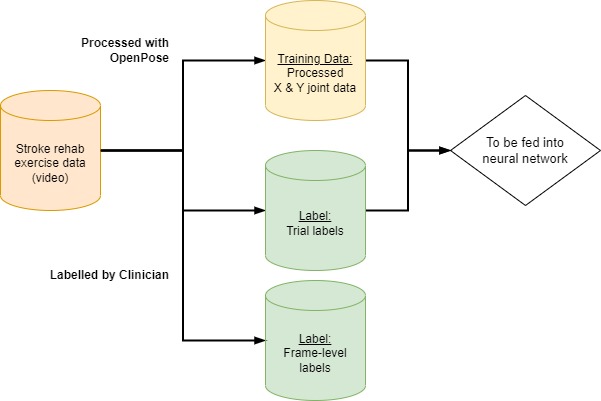}
    \caption{Data Preprocessing Diagram}
    \label{fig:data-preprocessing}
\end{figure}

\begin{figure}[h]
    \centering
    \includegraphics[scale=0.5]{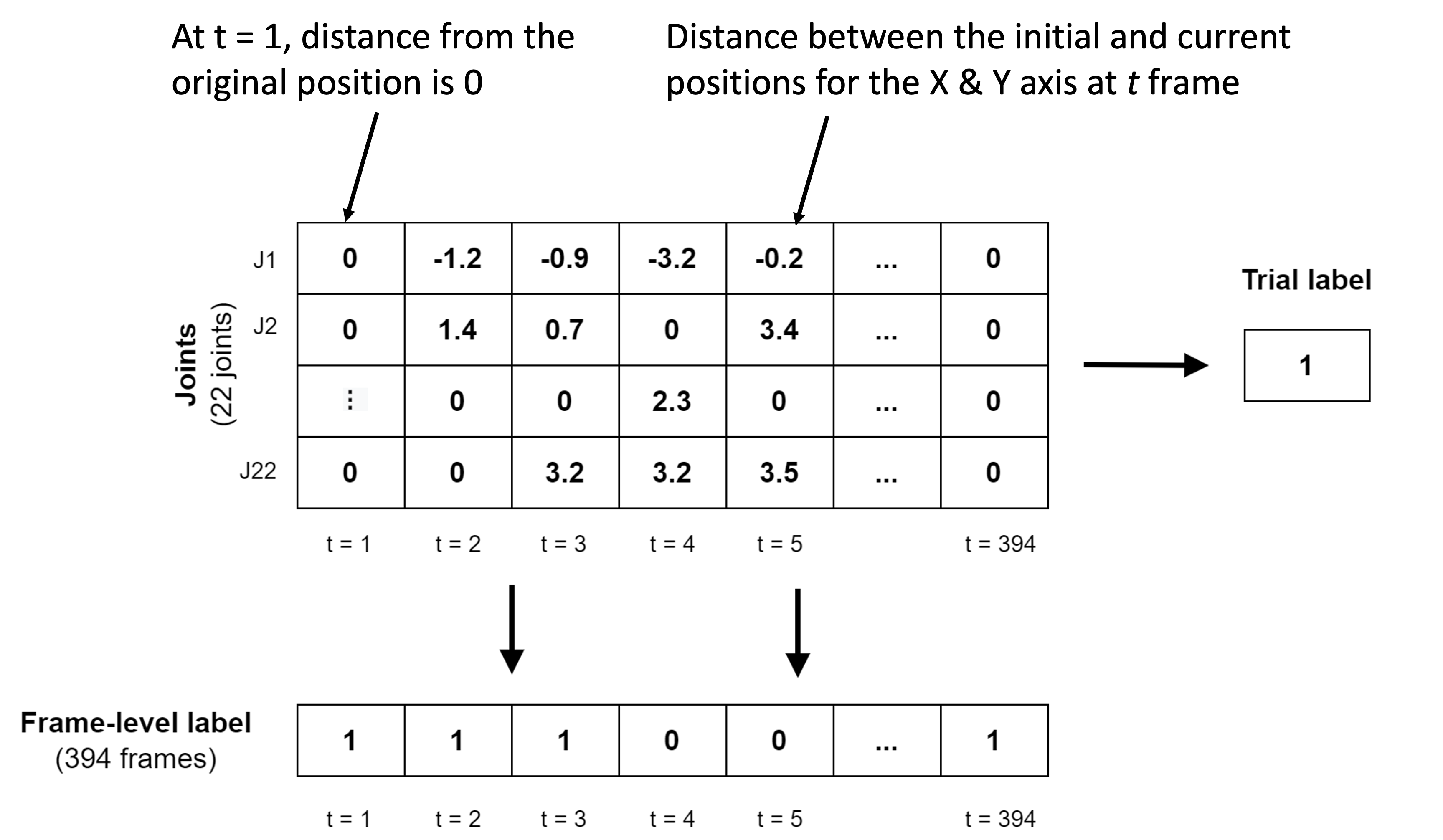}
    \caption{Shape of Data and Label after Preprocessing for an Exercise Trial}
    \label{fig:data-shape}
\end{figure}

\begin{figure}[h]
    \centering
    \includegraphics[scale=0.9]{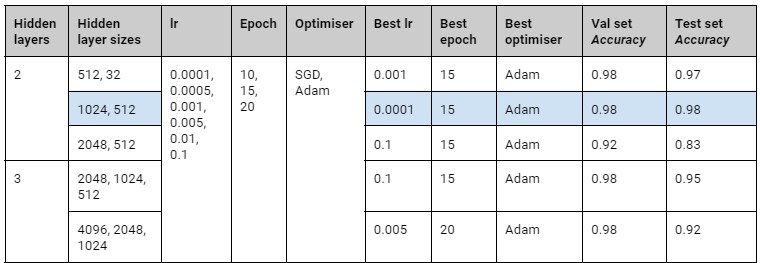}
    \caption{Summary of Model Parameters and Performance. The best performed architecture and parameters are higlighted in blue.}
    \label{fig:model-summary}
\end{figure}

\begin{figure}[h]
    \centering
    \includegraphics[scale=0.4]
    {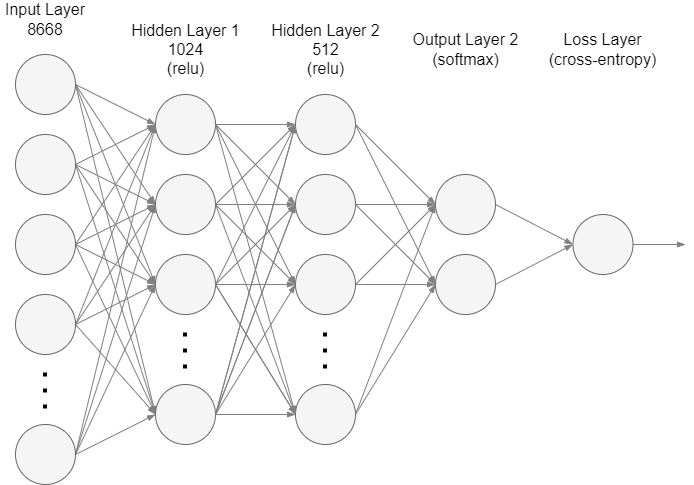}
    \caption{Final Model Architecture}
    \label{fig:final-model}
\end{figure}

\begin{figure}[h]
    \centering
    \includegraphics[scale=0.7]{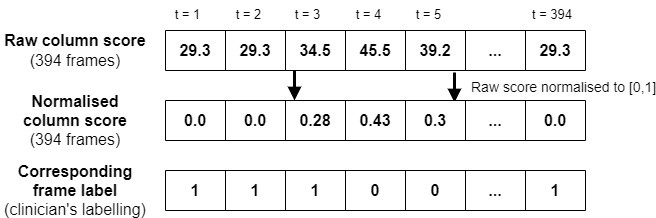}
    \caption{Score Computation Process}
    \label{fig:score-compute}
\end{figure}

\begin{table}[h]
    \centering
    \renewcommand{\arraystretch}{1.2}
    \begin{tabular}{|l|c|c|c|}
        \hline
        Experiment & Group 0 Scores & Group 1 Scores & Total \\ 
         \hline
         (1) All trials & 11,396 (9.64\%) & 106,804 (90.36\%) & 118,200 \\ 
         \hline
         (2) All trials without padded frames & 11,396 (9.64\%) & 37,048 (76.48\%) & 48,444 \\
         \hline
         (3) Compensatory trials without padded frames & 8,602 (65.75\%) & 4,485 (34.27\%) & 13,087 \\
         \hline
         (4) Window size = 5 (best performance) & 1,722 (65.25\%) & 917 (34.75\%) & 2,639 \\
         \hline
    \end{tabular}
    \caption{Data Distribution}
    \label{tab:experiment-distri}
\end{table}

\begin{figure}[h]
    \centering
    \includegraphics[scale=1.0]{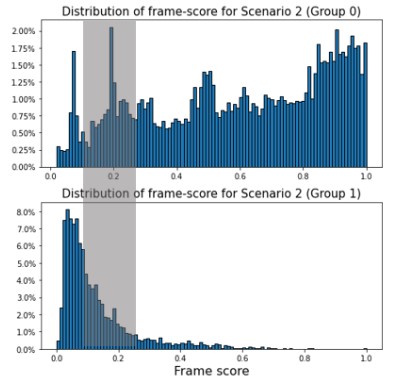}
    \caption{The histogram of normalized gradient scores using normal and abnormal, compensatory frames. The grey area indicates the overlapping areas between normal and abnormal frames}
    \label{fig:grey-area}
\end{figure}

\end{document}

%% file: math_commands.tex
\usepackage{amsmath,amsfonts,bm}
\usepackage{comment}
\usepackage{soul}
\usepackage{caption,subcaption,graphicx}

\def\eqref#1{equation~\ref{#1}}

\def\1{\bm{1}}

\DeclareMathAlphabet{\mathsfit}{\encodingdefault}{\sfdefault}{m}{sl}
\SetMathAlphabet{\mathsfit}{bold}{\encodingdefault}{\sfdefault}{bx}{n}



%% file: iclr2023/contents/intro.tex
\section{Introduction}
Artificial intelligence (AI) models are increasingly being explored to support high-stake decision-making, such as industrial predictive maintenance \citep{mobley2002introduction} and healthcare (e.g. heart-beat anomaly detection \citep{chuah2007ecg}, cancer diagnosis \cite{mckinney2020international}, rehabilitation assessment \cite{lee2021human}). A misprediction by AI models in these high-stake domains could have serious and lasting impacts on people and communities. As the end-users are unwilling to adopt these technologies without explaining AI model outcomes regardless of its performance \citep{wang2020should}, there is a growing emphasis to explain an AI model to end-users for their informed decision making \cite{cai2019human,lee2021human,rojat2021explainable}. 

There has been an expanding focus and research in explainable AI (XAI) methods for image and tabular data \citep{rojat2021explainable}. For instance, post-hoc explanations generate approximations of an AI model outcome by producing understandable information or explore saliency maps of an AI model outcome for image data \cite{simonyan2013deep}. Specifically, these saliency maps calculate the gradient of the loss function with respect to the input pixels for the class of an AI model outcome. Thus, these maps are useful to provide insights into which part of an image is considered important for the model outputs.

While saliency maps are traditionally for image data, previous studies have explored adapting saliency map methods for time series data \citep{rojat2021explainable}. \citet{goodfellow2018towards} used Class Activation Maps (CAM), a variant of saliency maps, \citep{zhou2016learning} to explain classifications made by a convolutional neural network (CNN) on univariate electrocardiogram (ECG) data. \citet{assaf2019explainable} used a CNN to forecast the average power consumption of a photo-voltaic power plant. This study utilized a more versatile form of CAM, Grad-CAM \citep{selvaraju2017grad} to explain the predictions from the CNN model. However, these previous works focus on exploring saliency maps as an explanation tool for a model’s output through qualitative evaluations. For instance, after generating sequences of saliency maps of a video, these work overlaid these maps on the input data and checked qualitatively whether the highlighted areas of time-series data are relevant to a model output or not \cite{samek2017explainable}. It remains unclear how well saliency maps can be used to identify salient frames in time-series data.

In this paper, we present a threshold score method that utilizes a weakly supervised machine learning (ML) model and a gradient-based explainable AI technique, saliency map to explain time-series data by identifying salient frames of the data. Using the dataset of three stroke rehabilitation exercises from 15 post-stroke survivors and labels on whether an exercise involves compensatory motions or not, we trained a feed-forward neural network model. Given the feed-forward neural network model, we computed the gradients of the loss function with respect to input data over all frames and normalized these gradient scores to determine whether a compensatory motion has occurred in a specific frame or not. After implementing a threshold score method, we utilized frame-level annotations on the presence of a compensatory motion to evaluate the feasibility of our method. Our results showed that our method can detect salient frames of a frame-level compensatory motion with a recall of 0.96 and an F2-score of 0.91. This work contributed to an empirical study on how well a weakly supervised ML model with a gradient-based explainable AI can be utilized as explainable AI techniques for time-series data (e.g. identifying salient frames of data).

%% file: iclr2023/contents/data_preprocessing.tex
\section{Physical Stroke Rehabilitation \& Data Preprocessing}
Post-stroke survivors require conducting physical stroke rehabilitation exercises to regain their functional abilities. While conducting exercises, they often compensate (e.g. lean their trunk to the side - Figure \ref{fig:sample_affected_comp}) for their limited functional abilities, which may be detrimental to long-term recovery \citep{alaverdashvili2008learned, pain2015effect}. Physical therapists monitor and correct compensatory movement in rehabilitation exercises, which are often recorded as patients perform them at home. However, examining through a video can be a long and tedious process. To address this challenge, this work explores a threshold-based method that utilizes a weakly supervised ML model and saliency maps to detect salient frames of time-series data (e.g. compensatory periods during an exercise).

This work utilizes the dataset that consists of 300 exercise trials performed by 15 patients \citep{lee2019learning}. Each patient performed 10 trials each with his/her affected and unaffected side. The affected side refers to the side of the body affected by the stroke, and vice versa for the unaffected side (Figure \ref{fig:movement-ex}). Given patients' exercise motions, this dataset includes two types of labels: (1) labels on each trial and (2) labels on each frame that indicates whether a compensatory motion occurs (0) or not (1 - normal). 

For processing the videos of patients performing an exercise, we utilized the Openpose\citep{8765346} to estimate the body key points of patients from the videos and extracted features that indicate the distance from an initial to a current position at every frame of patients' exercise videos. We identified the maximum frame number of the dataset (i.e. 394) and padded an initial position to samples that have a smaller number of frames than the maximum frame number for downstream model training (Appendix. Figure \ref{fig:data-preprocessing} and Figure \ref{fig:data-shape}). 

%% file: iclr2023/contents/model_training.tex
\section{Gradient-based Explainable Technique for Time-Series Data}

We describe the implementations of a weakly supervised machine learning model, a gradient-based explainable AI technique (i.e. saliency map), and a threshold-based method to explain time-series data (i.e. identifying salient frames of data) in the following sections. 

\subsection{A feed-forward neural network Model}
We applied an 80-20 split on the dataset to obtain the training (i.e. 240 exercises) and test (i.e. 60 exercises) data sets to train a feed-forward neural network model to detect whether a compensatory motion occurs in an exercise or not. Using Skorch \citep{skorch} and three-fold cross-validation, we grid-searched different architectures and parameters and implemented a feed-forward neural network model with a test accuracy of 0.98. We summarize the architectures, hyper-parameters, and performances of the models in the Appendix. Figure \ref{fig:model-summary} and describe the final, best-performing architecture in the Appendix. Figure \ref{fig:final-model}.

\subsection{Computing a Frame Score from Saliency Map}\label{sect:frame-score}
After implementing a feed-forward neural network, we computed gradients of the loss function with respect to input data over all frames. We then aggregated gradient scores of features at each frame level and normalized this frame-level score into [0, 1] (Appendix. Figure \ref{fig:score-compute}). 

We hypothesize that there is a threshold value between the score range [0,1] where scores less than or equal to the threshold score indicates normal movement, and scores more than the threshold score indicates compensation. Frames involving compensation will be deemed more important, and thus result in a higher gradient score for the time frame. 

%% file: iclr2023/contents/understand_saliencymap.tex
\section{Experiments}
We conducted both qualitative and quantitative evaluations to explore the feasibility of a gradient-based explainable technique for time-series data. For the qualitative evaluation, we utilized a sample exercise and checked whether the most important frames and features of the exercise using a saliency map can provide relevant insights into a compensatory motion (Figure \ref{fig:sample_affected_comp}).

In addition, we conducted quantitative evaluations of how well our threshold-based method can detect compensatory motions at the frame level. We explored the threshold value ranging from 0 to 1 and empirically selected an optimal threshold value. For the evaluation of our threshold-based approach, we utilized frame-level annotations and computed recall and f2-score with the $\beta = 2$. As our dataset is heavily skewed (e.g. containing a larger number of normal frames than abnormal frames with compensatory motions), we explored the effect of (i) removing zero-padded normal frames and (ii) utilizing only samples that include compensatory movements. Also, we explored the effect of aggregating a score over multiple windows ($5, 10, 15, 20$). For post-analysis, the scores have been sorted into 2 groups: Group 0 contains scores where their corresponding frame label = 0, while Group 1 contains scores where their frame label = 1. The data distributions of Group 0 and 1 are listed in Appendix. Table \ref{tab:experiment-distri}.

%% file: iclr2023/contents/analysing_score.tex
\section{Results}

\subsection{Qualitative Evaluation}
\begin{figure}[tp!]
\centering
\begin{subfigure}[t]{.19\columnwidth}
\centering
  \includegraphics[width=0.56\columnwidth]{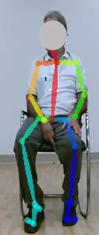}
  \caption{}
  \label{fig:sample_affected_initial}
\end{subfigure}
\begin{subfigure}[t]{.19\columnwidth}
\centering
  \includegraphics[width=0.8\columnwidth]{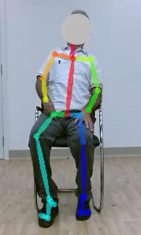}
  \caption{}
  \label{fig:sample_affected_comp}
\end{subfigure}
\begin{subfigure}[t]{.6\columnwidth}
\centering
  \includegraphics[width=1.0\columnwidth]{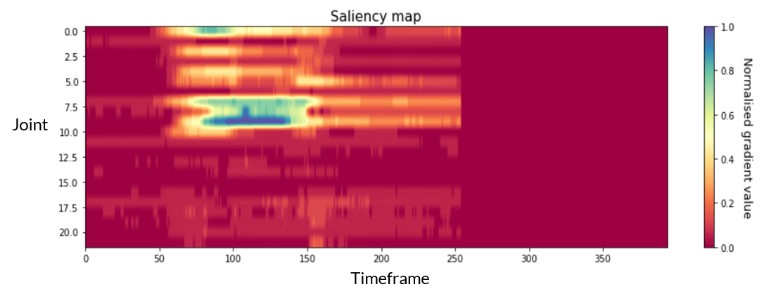}
  \caption{}
  \label{fig:saliency-map}
\end{subfigure}
\caption{(a) The initial position and (b) compensatory motion of a post-stroke survivor. (c) the saliency map of a corresponding post-stroke survivor's exercise}~\label{fig:sample-exercise-saliencymap}
\end{figure}

Figure \ref{fig:saliency-map} shows a saliency map for a post-stroke survivor's exercise trial (Figure \ref{fig:sample_affected_initial} and \ref{fig:sample_affected_comp}). Blue-green regions denote the frames and features with the highest importance, while orange-red regions indicate those with the least importance. The flat red region at the end of the map indicates zero-padded frames. For this exercise trial, our method identified 50 - 60 frames and joint features (0: HeadX, 2: NeckX, 5: ShoulderRightY, 6: ElbowRightX, 7: ElbowRightY, 8: WristRightX, 9: WristRightY, and 10:ShoulderLeftX) are most important with respect to the model classification. After qualitatively reviewing the corresponding compensatory frames (Figure \ref{fig:sample_affected_comp}), we found the potential of our method to provide an insight into a compensatory motion: When a post-stroke survivor strives to raise his right wrist to complete an exercise, he involved compensatory motion by moving his head and left shoulder to the side and right shoulder upward.

\subsection{Quantitative Evaluations}
Table \ref{tab:experiment-results} describes the quantitative evaluations of how well our approach can identify salient frames with compensatory motions.
We found that using only compensatory trials without padded frame scores gave the best results of recall = 0.96, F2-score = 0.91. As we utilized the dataset of more balanced distributions from all trials (90\% of normal frames and 9\% of compensatory frames in Appendix. Table \ref{tab:experiment-distri}) to compensatory trials without padded frames (34\% of normal frames and 65 of compensatory frames in Appendix. Table \ref{tab:experiment-distri}), our approach significantly improved its performance to identify frame-level compensatory motions. When it comes to the computation of a gradient with different window sizes (i.e. 5, 10, 15, 20), we did not find any significantly different results. 
We achieved the best performance to detect frame-level compensatory motions by using only scores from samples involving compensatory motions and removing padded frame scores. 

\begin{table}[ht]
    \centering
    \renewcommand{\arraystretch}{1.2}
    \begin{tabular}{|l|c|c|c|}
        \hline
        Experiment & Best Threshold Score & Recall & F2-Score \\ 
         \hline
         (1) All trials & 0.4 & 0.44 & 0.44 \\ 
         \hline
         (2) All trials without padded frames & 0.4 & 0.44 & 0.44 \\
         \hline
         (3) \textbf{Compensatory trials without padded frames} & \textbf{0.36} & \textbf{0.96} & \textbf{0.91} \\
         \hline
    \end{tabular}
    \caption{Experiment results using datasets with different data distributions of normal and compensatory frames and window size of 5. The best result is boldfaced}
    \label{tab:experiment-results}
\end{table}

\subsection{Discussion}
In this work, we contributed to an empirical study that explores the feasibility of using a weakly supervised ML model and a gradient-based explainable AI (XAI) technique, saliency map for explaining time-series data (e.g. post-stroke rehabilitation exercises). Our qualitative and quantitative results are aligned with the hypothesis discussed in Section \ref{sect:frame-score}. Specifically, our results demonstrate the potential of our approach to detect frame-level compensatory motions of post-stroke survivors with a recall of 0.96 and an F2-score of 0.91. Our approach enables a therapist to pinpoint an important time period of a video that the therapist should prioritize reviewing and support human-AI collaborative annotations instead of a time-consuming manual labeling process \cite{lee2022towards}. However, there are several limitations to be addressed. 

First, our quantitative evaluations using datasets with different distributions between normal and abnormal frames suggest the importance of utilizing a balanced dataset for applying our threshold-based method using a gradient-based XAI technique. When we utilized the unbalanced dataset with both unaffected and affected motions with padded frames (9\% of compensatory frames and 90\% of normal frames - Appendix. Table \ref{tab:experiment-distri}), we found that some scores were normalized to higher values and led to higher misdetections of frame-level compensatory motions. 

In addition, it is also important to consider a more systematic approach to dynamically determine an \textit{``ideal''} threshold value and address overlapping regions of scores between normal and compensatory frames (Appendix. Figure \ref{fig:grey-area}). Also, as this work only explores a feed-forward neural network model and a saliency map to compute gradient scores using video data from 15 post-stroke survivors, it is necessary to conduct additional studies to demonstrate the generalizability of our approach to other ML algorithms, gradient-based explainable AI methods, and data modalities (e.g. text).